# Patient Clustering Improves Efficiency of Federated Machine Learning to predict mortality and hospital stay time using distributed Electronic Medical Records

Li Huang[1] and Dianbo Liu[2,3,*]

1. Academy of Arts and Design, Tsinghua University, Beijing, China, 10084
2. Computer Science & Artificial Intelligence Laboratory, Massachusetts Institute of Technology, Cambridge , MA ,USA, 02139
3. Medical School, Harvard University, 25 Shattuck St, Boston, MA 02115

*. Corresponds to: dianbo@mit.edu

**Abstract**

Electronic medical records (EMRs) supports the development of machine learning algorithms for predicting disease incidence, patient response to treatment, and other healthcare events. But insofar most algorithms have been centralized, taking little account of the decentralized, non-identically independently distributed (non-IID), and privacy-sensitive characteristics of EMRs that can complicate data collection, sharing and learning. To address this challenge, we introduced a community-based federated machine learning (CBFL) algorithm and evaluated it on non-IID ICU EMRs. Our algorithm clustered the distributed data into clinically meaningful communities that captured similar diagnoses and geological locations, and learnt one model for each community. Throughout the learning process, the data was kept local on hospitals, while locally-computed results were aggregated on a server. Evaluation results show that CBFL outperformed the baseline FL algorithm in terms of Area Under the Receiver

Operating Characteristic Curve (ROC AUC), Area Under the Precision-Recall Curve (PR AUC), and communication cost between hospitals and the server. Furthermore, communities' performance difference could be explained by how dissimilar one community was to others.

## 1. Introduction

That EMRs improve the quality of healthcare has been endorsed by various evidences including enhanced performance of patients with chronic illness [1-5], reducing unnecessary medical examinations [6], cost saving for healthcare providers [7], better medical education [8] and more. To reap the benefits to a larger extent, machine learning applications have been developed on EMRs: for instance, ensemble learning of regression, k-nearest neighbor, decision trees and support vector machines for predicting type 2 diabetes (T2D) one year prior to diagnosis of diabetes [9], prediction of suicide risk via EMR-driven nonnegative restricted Boltzmann machines [10], classification of normal versus age-related macular degeneration OCT images using deep neural networks [11], and modeling of hospital readmission rates by a multistep Naïve Bayes-based learning strategy [12].

While such applications demonstrated promising perspectives towards translation of EMRs into improved human health [13], nevertheless they were developed under the premise that EMRs could be easily shared across silos and stored in centralized data warehouses. Generated by individual patients and in diverse hospitals/clinics, EMRs are distributed and sensitive in nature. This may impede adoption of machine learning on EMRs in reality, and has entailed researchers to raise concerns on central storage of EMRs and on security, cost-effectiveness, privacy and availability of medical data sharing [14-24]. These concerns can be addressed by federated machine learning (FL) that keeps both data and computation local in distributed silos and aggregates locally computational results to train a global predictive model [25]. Indeed, FL precludes the need of data collection and sharing, and thus can serve as a desirable framework for developing machine learning applications on privacy-sensitive EMRs.

However, FL may underperform when data is non-identically independently distributed [25-28], as EMRs usually are [29]. To tackle this non-IID challenge and inspired by deep embedding clustering [30], we proposed a community-based federated learning (CBFL) algorithm that clustered EMR data into several communities and simultaneously trained one model per community, so that the learning process became markedly more efficient than FL. Success of data clustering (albeit being centralized analyses) has been reported in previous medical studies such as quality assessment of diabetes physician groups [31], identification of cancer symptom clusters to benefit therapeutics [32], and delineation of chronic pain patient subgroups for improving treatment [33]. In this study, by presenting the development and evaluation of CBFL, we demonstrate the application of decentralized clustering together with federated machine learning to make predictions on ICU EMRs.

## 2. Materials and Methods
### 2.1. eICU data
CBFL was developed based on the eICU collaborative research database [34], which contains highly granular critical care data of 200,859 patients admitted to 208 hospitals from across the United States. Our study mainly concerned with three dimensions:

- drugs administered on patients during the first 48 hours of ICU stay (1,399 binary *drug features* in total)
- unit discharge status that specifies patients' condition upon leaving ICU (*mortality*, 0 for alive and 1 for expired)
- unit discharge offset that records the number of minutes from unit admission to discharge (ICU *stay time*, with an average of 3,858 minutes)

Extracting these dimensions out of the database yielded a smaller dataset of 126,490 patients coming from 58 hospitals. Furthermore, we selected 50 hospitals whose patient count was over 600 and, from each of them, randomly sampled 560 patients to form the final dataset of 280,000 examples. Results of cohort analysis on this data will be presented in Section 3.1.

## 2.2. CBFL

Algorithm 1 displays the three procedures involved in CBFL. During encoder training, each client (that is, hospital) learnt a denoising autoencoder $f_{autoencoder}$ initialized with $w_{0,autoencoder}$ for $E_1$ epochs and returned only the trained weights of encoder $w_{1,encoder}^c$ to the server for average. $N$, $n^c$, and $f_{encoder}$ denoted the total number of examples, the size of each client, and the averaged encoder, respectively. During k-means clustering, each client used $f_{encoder}$ to transform its data into representations $X^c$ and sent the average representation $\overline{X^c}$ to the sever. Then, the server learnt a k-means clustering model $f_{kmeans}$ with $K$ centroids (that is, communities) on $\overline{X^c}$s from all clients. During community-based learning, the server initialized a series of $K$ neural network models $f_1, f_2, ..., f_K$ with the same weights $w_0$; each client received all $K$ models from the server and learnt each model on its full data for $E_2$. Meanwhile, $f_{encoder}$ and $f_{kmeans}$ were used to determine which cluster each example belonged to. The size of clusters was denoted $m_1^c, m_2^c, ..., m_K^c$ and returned together with the learnt weights $w_K^c$ to the server, where each model was updated by taking the weighted average of $w_K^c$s based on $m_1^c, m_2^c, ..., m_K^c$. The updated models were sent to each client for the next round of training. This community-based learning process was repeated until the algorithm converged. Given a test example, CBFL would firstly convert its features into encodings by $f_{encoder}$, then define its community by $f_{kmeans}$ and finally use the corresponding community model to make prediction. These three procedures of CBFL are visualized in Figure 1.

```
Algorithm 1 Community-Based Federated Learning
 1: procedure ENCODER TRAINING(C, E_1)                    ▷ First procedure
 2:     initialize weights w_{0,autoencoder} of denoising autoencoder f_{autoencoder}
 3:     for each client c = 1, 2, ..., C in parallel do
 4:         train f_{autoencoder} for E_1 epochs to obtain w^c_{1,encoder}
 5:         return w^c_{1,encoder} to server        ▷ return weights of encoder layers only
 6:     w_{1,encoder} ← ∑_{c=1}^{C} (n^c/N) w^c_{1,encoder}  ▷ perform one update to obtain f_{encoder}
 7: procedure K-MEANS CLUSTERING(K, C)                    ▷ Second procedure
 8:     for each client c = 1, 2, ..., C in parallel do
 9:         use f_{encoder} to obtain encoded features X^c of each example
10:         return X̄^c = ∑_{i=1}^{n^c} (X^c / n_c) to server   ▷ return average features
11:     initialize K cluster centroids from {X̄^1, X̄^2, ..., X̄^C}
12:     train k-means clustering model f_{kmeans}
13: procedure COMMUNITY-BASED LEARNING(K, C, E_2)         ▷ Third procedure
14:     initialize K community NN models {f_1, f_2, ..., f_K} with same weights w_0
15:     while not converged do
16:         for k in 1...K in parallel do         ▷ simultaneously train K models
17:             for each client c = 1, 2, ..., C in parallel do
18:                 train f^c_k for E_2 epochs to obtain w^c_k
19:                 use f_{encoder} and f_{kmeans} to determine cluster of each example
20:                 count examples in each cluster {m^c_1, m^c_2, ..., m^c_K}
21:                 return w^c_k, {m^c_1, m^c_2, ..., m^c_K} to server
            w_k ← (∑_{c=1}^{C} m^c_k w^c_k) / (∑_{c=1}^{C} m^c_k)   ▷ update weights of each community model
22:     end while
```

## 2.3. parameter set of CBFL

The autoencoder $f_{autoencoder}$ had a structure of five fully connected hidden layers with 200, 100, 50, 100 and 200 units, respectively, using the rectified linear unit (ReLu) activation function. The output layer used the sigmoid function because of binary input *drug features*. The number of epochs $E_1$ was set to five, meaning that each hospital would train $f_{autoencoder}$ on every example for five times. We chose Adaptive Moment Estimation (Adam) as the stochastic optimizer with default parameters (the learning rate $\eta = 0.001$ and the exponential decay rates for the moment estimates $\beta_1 = 0.9$ and $\beta_2 = 0.999$) on the categorical cross-entropy loss function. For the *k*-means model $f_{kmeans}$, we used various numbers of centroids (five, 10, 15 and the extreme case of one centroid per hospital) to evaluate CBFL. Each one of the community models $f_1, f_2, ..., f_K$ consisted of three hidden layers with 20, 10 and five hidden units respectively and activated by ReLu. Same as $f_{autoencoder}$, we used sigmoid as the output layer activation function and Adam with default setting as the optimizer. The loss function was given by binary cross-entropy. The baseline FL model was constructed with the same parameters as the community models.

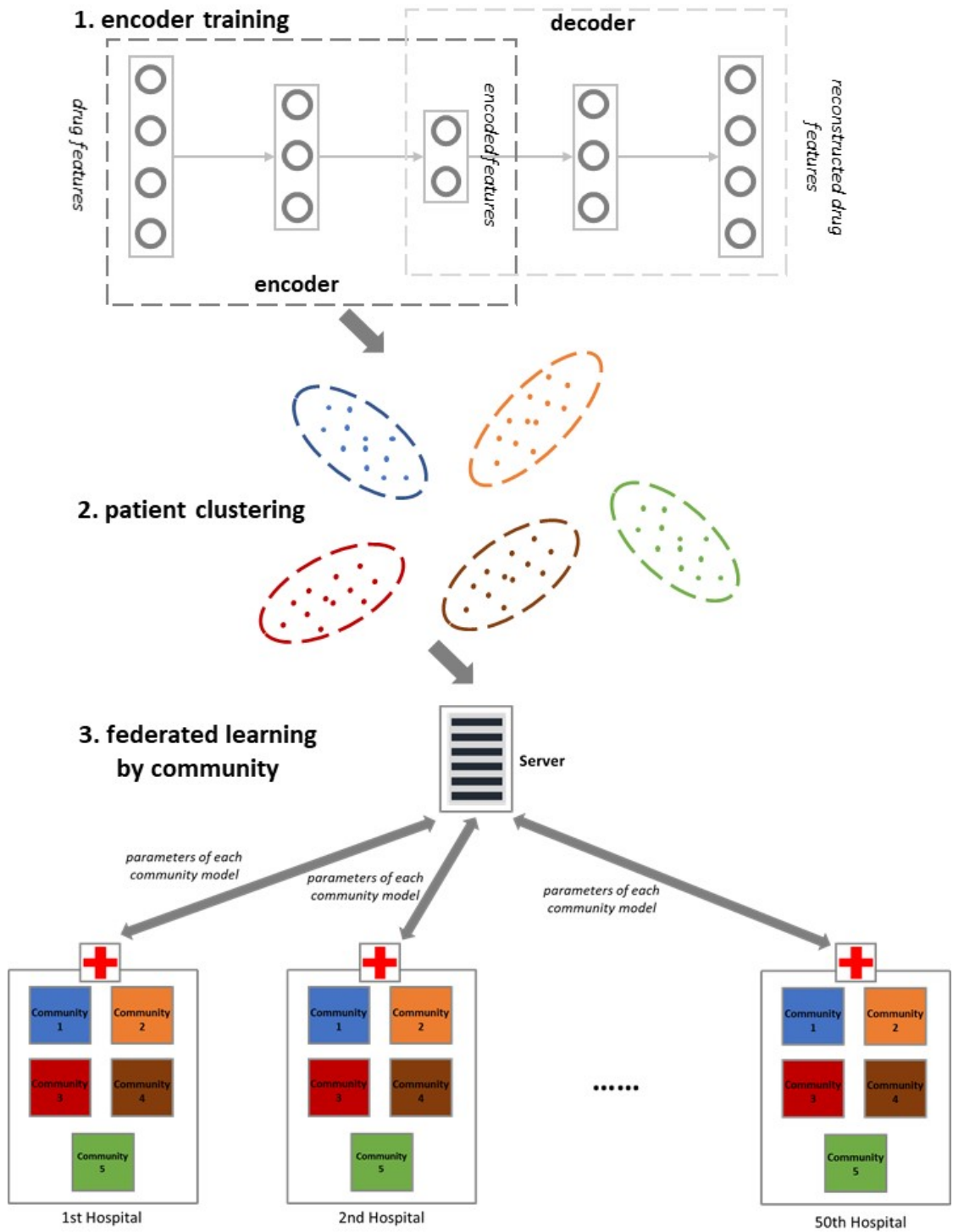

Figure 1: flowchart of CBFL. A denoising autoencoder was trained on each hospital's data and averaged at the server. Subsequently, encoder layers were used to convert patients' *drug features* into privacy-preserving representations that were in turn used for patient clustering by *k*-means. In this figure, patients were clustered into five communities as an example. Lastly, each hospital individually learnt five

community models and sent them to the server for weighted average based on community size.

## 3. Results

### 3.1. Cohort analysis

This study involved EMRs from 50 hospitals, each containing 560 critical care patients. Table 1 summarizes the study cohort's information, including patient count, gender, age, mortality/survival rate, prolonged stay time rate, and the most frequent diagnosed diseases for ICU patients. The cohort contained more males than females (54.95% versus 45.03%) and most patients (61.82%) were aged above 60 years old. The mortality rate was 4.98%, and 6.12% of the study cohort experienced a prolonged unit stay time. In our study, patients had a prolonged unit stay time if they experienced greater than or equal to eight days of stay, and non-prolonged otherwise. As for the top frequent diagnoses, patients sent to ICU most likely suffered from diseases related to burns/trauma, cardiovascular, endocrine, gastrointestinal, other general conditions, hematology, infectious diseases, musculoskeletal, neurologic, obstetrics/gynecology, and oncology.

|  | count | percentage |
|---|---|---|
| patients | 28,000 | - |
| male | 15,386 | 54.95% |
| female | 12,609 | 45.03% |
| unknown gender | 5 | 0.02% |
| age≤20 | 267 | 0.95% |
| 20<age≤40 | 2,552 | 9.11% |
| 40<age≤60 | 7,943 | 28.37% |
| 60<age≤80 | 12,630 | 45.11% |
| 80<age | 4,680 | 16.71% |
| death | 1,395 | 4.98% |
| alive | 26,605 | 95.02% |
| patients with prolonged unit stay time | 1,713 | 6.12% |
| top 10 frequent diagnoses | • burns/trauma • cardiovascular • endocrine • gastrointestinal • general • hematology • infectious diseases • musculoskeletal • neurologic • obstetrics/gynecology • oncology | |

Table 1: cohort analysis of 28,000 patients from 50 hospitals

## 3.2. Community analysis

Patient clustering was a key step in our algorithm: since patients with similar features were grouped together, community-based learning (that is, learning an independent model on each community) would be easier than learning one whole model on all patients. To illustrate what common features were shared among patients in the same community, we clustered the 28,000 patients into five communities and carried out enrichment analysis of diagnoses in them. Table 2 lists the number of patients and overrepresented diagnoses with adjusted *p*-values within each community. It can be noted that every community exhibited a different focus: for instance, *Community* 1 tended to primarily capture neurologic, endocrine and burns/trauma diseases, whereas *Community* 3 concerned more with pulmonary, cardiovascular and gastrointestinal diseases.

|  | patient count | overrepresented diagnoses (with adjusted *p*-values) |
|---|---|---|
| Community 1 | 5,027 | - neurologic (1.10e-20)<br>- endocrine (4.95e-14)<br>- burns/trauma (1.11e-11)<br>- hematology (5.23e-09)<br>- infectious diseases (1.57e-06)<br>- renal (7.22e-06) |
| Community 2 | 6,726 | - cardiovascular (1.27e-29)<br>- transplant (0.00105)<br>- hematology (0.00667)<br>- oncology (0.0249) |
| Community 3 | 2,322 | - pulmonary (3.00e-26)<br>- cardiovascular (9.99e-14)<br>- gastrointestinal (1.19e-10) |
| Community 4 | 6,247 | - pulmonary (3.97e-28)<br>- cardiovascular (1.05e-25)<br>- toxicology (0.00201) |
| Community 5 | 7,678 | - endocrine (1.75e-24)<br>- burns/trauma (5.90e-24)<br>- hematology (9.77e-12)<br>- infectious diseases (9.81e-11)<br>- gastrointestinal (1.72e-10)<br>- toxicology (1.37e-06) |

|  |  | - oncology (4.08e-05)<br>- general (0.00347)<br>- transplant (0.0138)<br>- surgery (0.0172) |

Table 2: enrichment analysis of diagnoses in five patient communities

In addition to the above cohort and community analyses considering the characteristics of patients, we further performed clustering at hospital level to reveal distinction between hospital communities. Figure 2 visualizes the 50 hospitals (labeled with their eICU IDs) clustered into five communities on a PCA plot. Separation between communities can be easily recognized, and *Communities* 1 and 5 had a larger size than the rest three. Moreover, geological bias could be found: *Community* 1 had 15 hospitals located in the Midwest (nine), the South (five) and the West (one) of the United States; *Community* 2 had seven hospitals, all situated in the South; *Community* 3 had eight hospitals, seven of which came from the West and one with unknown location; *Community* 4 had three Midwestern hospitals and two Western hospitals; *Community* 5 had 15 hospitals, one with unknown location and the others residing in the Northeast (three), the Midwest (five), the South (four), and the West (two). In summary, *Community* 2 seemed to capture Southern hospitals only and *Community* 3 tended to accommodate hospitals from the West, while no notable bias was observed in the other communities. Supplementary Table 1 contains full information of each hospital.

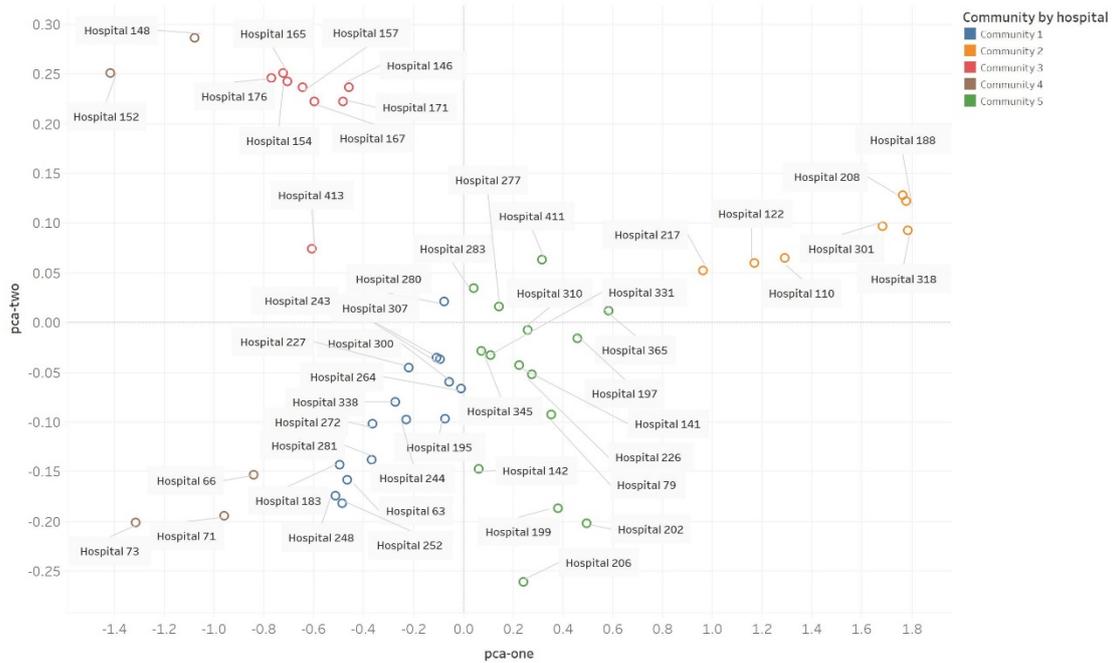

Figure 2: communities by hospital (50 hospitals clustered into five communities and visualized on a PCA scatterplot)

### 3.3. Mortality prediction

#### 3.3.1. same hospitals in training and test sets

*Mortality* was predicted based on patients' prescribed *drug features*. The training dataset was formed by randomly selecting 400 patients from each of the 50 hospitals, and thus had a size of 20,000 examples; the test dataset contained the remaining 160 patients from each hospital, totaling 8,000 examples. All patients were labeled with their unit discharge status (1 for expired and 0 for alive). Evaluation metrics included not only predictive accuracy in terms of ROC AUC and PR AUC but also the number of communication rounds between the server and hospitals to complete the training process. The ROC curve was generated by plotting true positive rate (TPR) versus false positive rate (FPR), while the PR curve was produced by plotting positive predictive value (PPV) against TPR. In our study, ROC AUC referred to the probability that CBFL would rank a randomly chosen mortal patient higher than a randomly chosen survival one, while PR AUC indicated the average precision across the recall range between 0 and 1.

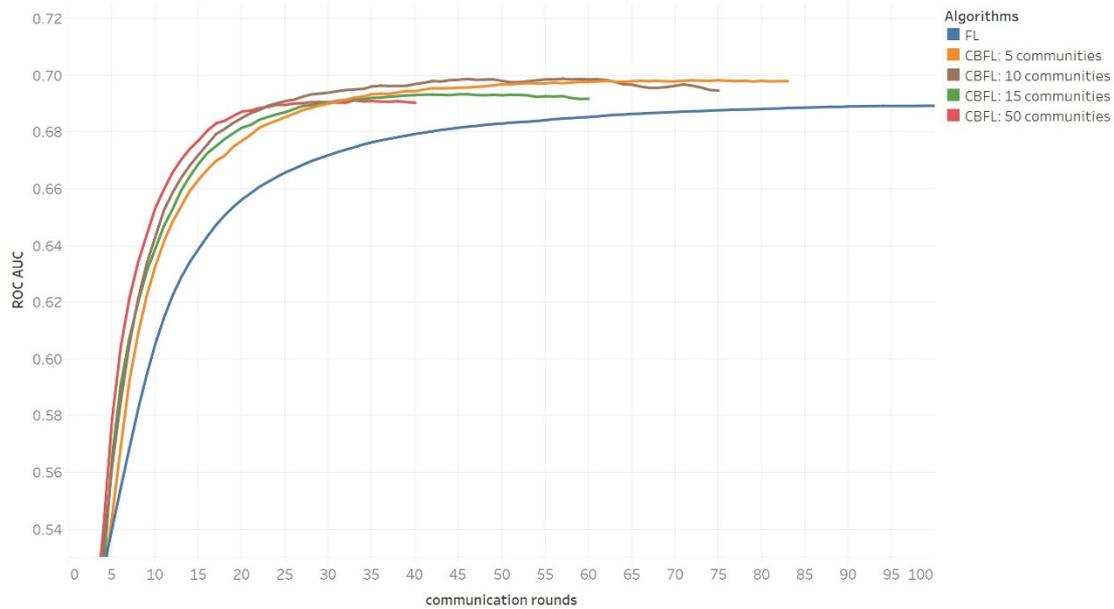

Figure 3: plot of ROC AUC against communication rounds in the *mortality* prediction task; FL and CBFL with 5, 10, 15 and 50 communities were compared; training and test data came from the same 50 hospitals

Figure 3 illustrates the curves of ROC AUC versus communication rounds for FL, CBFL with five communities, 10 communities, 15 communities and the extreme case of 50 communities (that is, one expected community per hospital). Two major messages are conveyed by the plots. First, CBFL consistently outperformed FL by converging to higher ROC AUCs with less communication rounds. FL achieved a final ROC AUC of 0.6895 and a PR AUC of 0.1107 in 101 rounds, whereas CBFL with five communities obtained a ROC AUC of 0.6984 and a PR AUC of 0.1430 in 75 rounds (see Table 3). Second, clustering patients into more communities tended to overfit CBFL, yielding slightly reduced ROC AUCs and PR AUCs, but meanwhile making the algorithm converge faster. For instance, when the number of communities was increased to 15, the ROC AUC, the PR AUC and communication rounds decreased to 0.6935, 0.1339 and 46, respectively. Neither FL nor CBFL performed better than centralized learning with a ROC AUC of 0.7368 and a PR AUC of 0.1449. This superiority of centralized learning was in line with reported in the literature of FL [25].

|  | ROC AUC | PR AUC | communication rounds |
|---|---|---|---|
| centralized learning | 0.7368 | 0.1449 | - |
| FL | 0.6895 | 0.1107 | 101 |
| CBFL: 5 communities | 0.6984 | 0.1430 | 75 |
| CBFL: 10 communities | 0.6989 | 0.1070 | 57 |
| CBFL: 15 communities | 0.6935 | 0.1339 | 46 |
| CBFL: 50 communities | 0.6913 | 0.1168 | 33 |

Table 3: summary of ROC AUCs, PR AUCs, and communication rounds after convergence in the *mortality* prediction task; centralized learning, FL and CBFL with 5, 10, 15 and 50 communities were compared; training and test data came from the same 50 hospitals

### 3.3.2. different hospitals in training and test sets

To evaluate the robustness of our model given different training/test data distributions, we prepared a training set of 19,600 examples from randomly chosen 35 hospitals and a test set of 8,400 from the remaining 15 hospitals. Unlike Section 3.3.1., no random split was performed in individual hospitals and patients in each hospital were used altogether; same as before, ROC AUC, PR AUC and communication rounds were used as evaluation metrics.

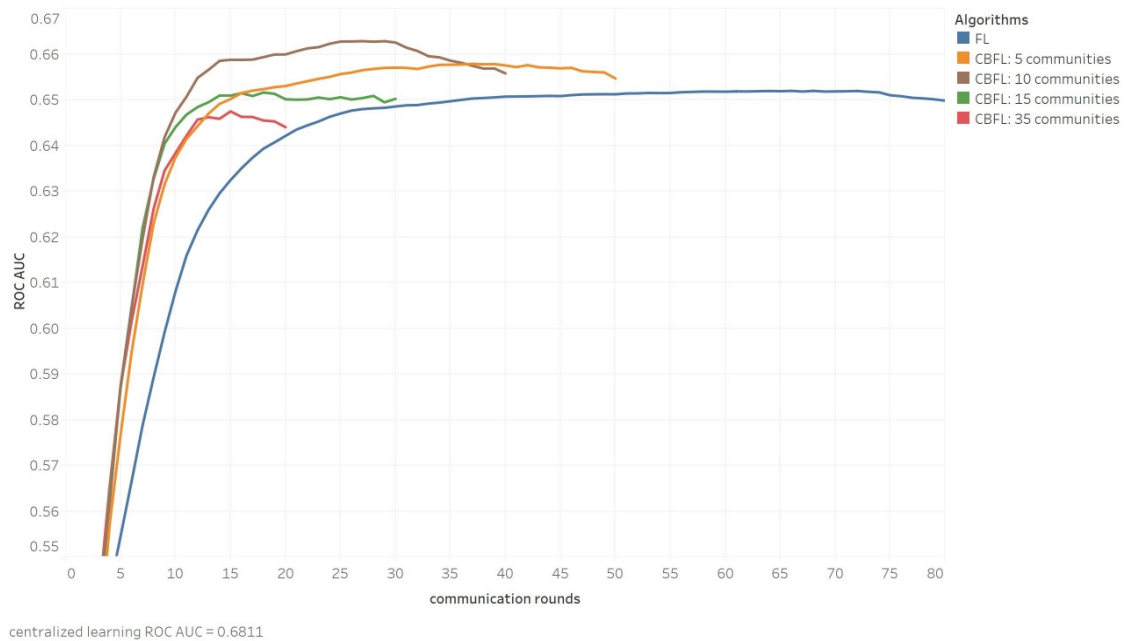

Figure 4: plot of ROC AUC against communication rounds in the *mortality* prediction task; FL and CBFL

with 5, 10, 15 and 35 communities were compared; training data came from randomly chosen 35 hospitals and test data from the remaining 15 ones

Figure 4 and Table 4 depict performance comparison between FL and CBFL with 5, 10, 15 and 35 communities. There was a drop in ROC AUC for centralizing learning (from 0.7368 to 0.6811), FL (from 0.6895 to 0.6520), and CBFL (from 0.69+ to 0.65+), resonating with the fact that inconsistent training/set data distributions lead to less effective learning. Nonetheless, both FL and CBFL converged faster, which we speculate resulted from fewer different hospitals (35, compared with 50 in previous evaluation) to learn in the training dataset. Specifically, FL reached its peak (a ROC AUC of 0.6520 and a PR AUC of 0.0871) in 66 communication rounds, and CBFL with 10 communities performed better than with any other community number, converging to a ROC AUC of 0.6628 and a PR AUC of 0.0912 in 27 rounds. In addition, the effect of overfitting was also observable: the more communities were clustered, the less ROC AUC would be achieved, albeit with fewer communication rounds.

|  | ROC AUC | PR AUC | communication rounds |
| --- | --- | --- | --- |
| centralized learning | 0.6811 | 0.0947 | - |
| FL | 0.6520 | 0.0871 | 66 |
| CBFL: 5 communities | 0.6579 | 0.0893 | 37 |
| CBFL: 10 communities | 0.6628 | 0.0912 | 27 |
| CBFL: 15 communities | 0.6516 | 0.1145 | 18 |
| CBFL: 35 communities | 0.6475 | 0.0920 | 15 |

Table 4: summary of ROC AUCs, PR AUCs, and communication rounds after convergence in the *mortality* prediction task; centralized learning, FL and CBFL with 5, 10, 15 and 35 communities were compared; training data came from randomly chosen 35 hospitals and test data from the remaining 15 ones

### 3.4. Stay time prediction

### 3.4.1. same hospitals in training and test sets

Like *mortality*, prediction of prolonged ICU *stay time* was based on *prescribed drug features*, with patients split in the same way as in Section 3.3.1. to form the training and test datasets, and assessed by the same evaluation metrics. Performance comparison of FL and CBFL is demonstrated in Figure 5 and Table 5. Here the AUC gap (0.02) was wider than that in *mortality* prediction task (0.01). FL achieved a ROC AUC of 0.6360 and a PR AUC of 0.0816 in 123 communication rounds, whereas the most performant CBFL with five communities obtained a ROC AUC of 0.6512 and a PR AUC of 0.0910 in 87 rounds. Again, the effect of overfitting became severer as the number of communities rose.

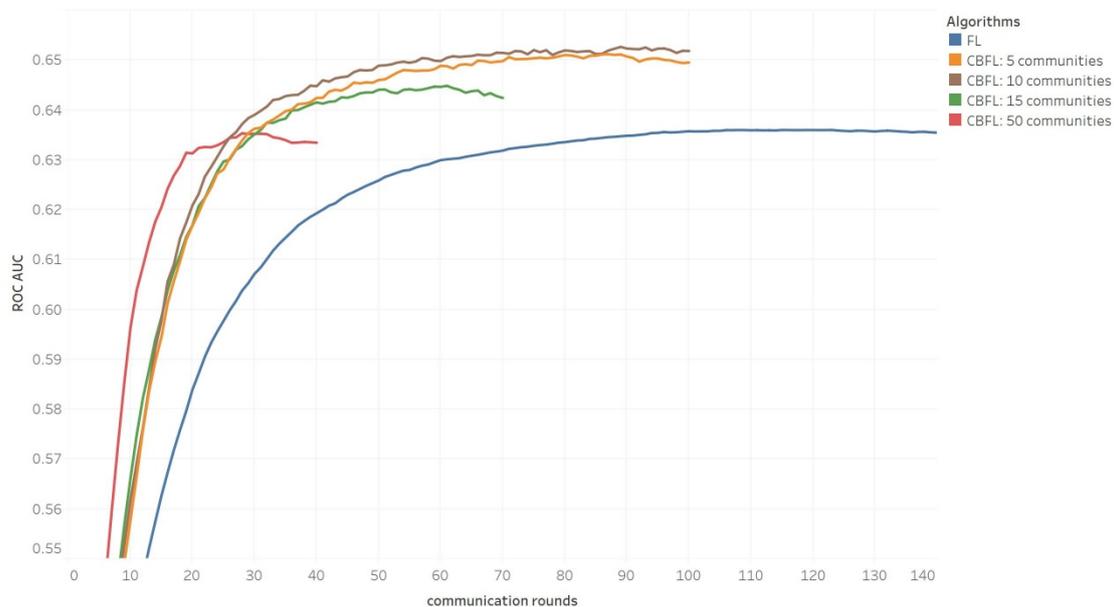

centralized learning ROC AUC = 0.7083

Figure 5: plot of ROC AUC against communication rounds in the *stay time* prediction task; FL and CBFL with 5, 10, 15 and 50 communities were compared; training and test data came from the same 50 hospitals

|  | ROC AUC | PR AUC | communication rounds |
|---|---|---|---|
| centralized learning | 0.7083 | 0.1145 | - |
| FL | 0.6360 | 0.0816 | 123 |
| CBFL: 5 communities | 0.6512 | 0.0910 | 87 |
| CBFL: 10 communities | 0.6527 | 0.0607 | 89 |
| CBFL: 15 communities | 0.6449 | 0.0549 | 61 |
| CBFL: 50 communities | 0.6353 | 0.0840 | 28 |

Table 5: summary of ROC AUCs, PR AUCs, and communication rounds after convergence in the *stay time* prediction task; centralized learning, FL and CBFL with 5, 10, 15 and 50 communities were compared; training and test data came from the same 50 hospitals

### 3.4.2. different hospitals in training and test sets

When the training and test datasets were prepared in the same manner as in Section 3.3.2. so that they came from different distributions, ROC AUCs reduced significantly from 0.7083 to 0.6189 for centralized learning, from 0.6360 to 0.6212 for FL, and 0.63+ to 0.62+ for CBFL, despite faster convergence (see Figure 6 and Table 6). It is worth noting that for the first time FL and CBFL outperformed centralized learning. We conjecture the reason to be that, in this particular task of predicting prolonged *stay time* on different training and test data, the beneficial effect of regularization [25] in federated machine learning outweighed the adverse effect of decentralization information loss. With five communities, CBFL exhibited the highest ROC AUC of 0.6400 and a PR AUC of 0.0822 in 23 communication rounds. The impact of overfitting on communication cost was less observable than that in previous sections, since raising the number of communities from five to 10 or 15 resulted in more communication rounds (31) rather than fewer.

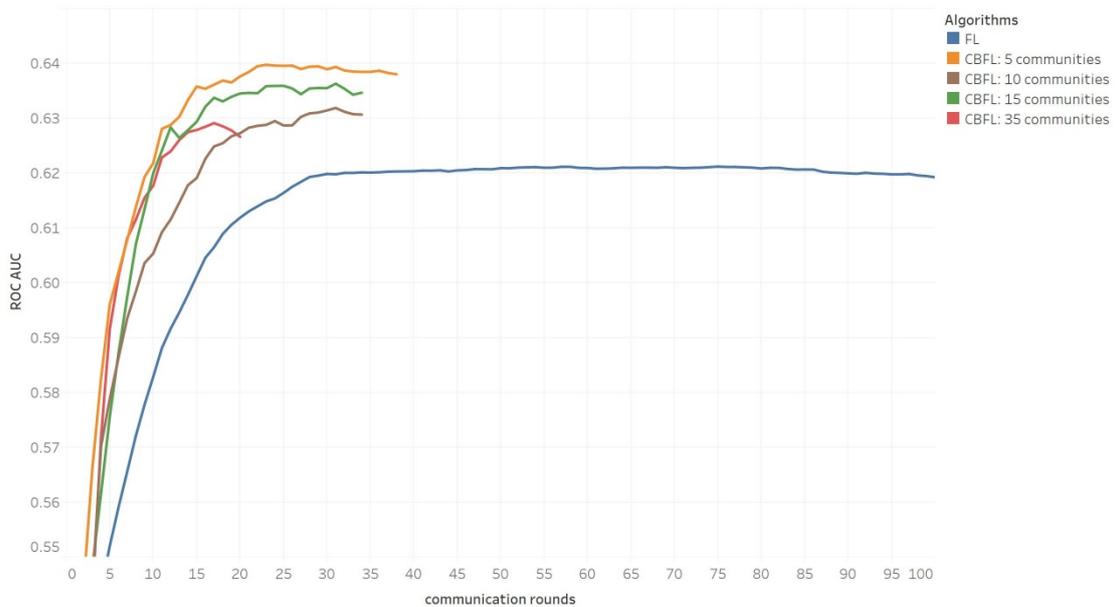

centralized learning ROC AUC = 0.6189

Figure 6: plot of ROC AUC against communication rounds in the *stay time* prediction task; FL and CBFL with 5, 10, 15 and 35 communities were compared; training data came from randomly chosen 35 hospitals and test data from the remaining 15 ones

|  | ROC AUC | PR AUC | communication rounds |
|---|---|---|---|
| centralized learning | 0.6189 | 0.0620 | - |
| FL | 0.6212 | 0.0720 | 75 |
| CBFL: 5 communities | 0.6400 | 0.0822 | 23 |
| CBFL: 10 communities | 0.6319 | 0.0786 | 31 |
| CBFL: 15 communities | 0.6364 | 0.0836 | 31 |
| CBFL: 35 communities | 0.6292 | 0.0694 | 17 |

Table 6: summary of ROC AUCs, PR AUCs, and communication rounds after convergence in the *stay time* prediction task; centralized learning, FL and CBFL with 5, 10, 15 and 35 communities were compared; training data came from randomly chosen 35 hospitals and test data from the remaining 15 ones

## 3.5. Community distribution analysis

The abovementioned evaluation results reveal that CBFL had better predictive accuracy in fewer communication rounds than FL in both mortality and stay time prediction tasks. Communities tended to accommodate patients of similar diagnoses and geological locations, making individual community models on average easier to learn than one model for all patients. In this section, we took CBFL with five communities for mortality prediction as an example to investigate and illustrate performance difference of each community model. As shown in Table 7, $Community\ 1$ exhibited the highest ROC AUC of 0.7561 and $Community\ 4$ yielded the highest PR AUC of 0.2155, while $Community\ 2$, the only one underperforming FL, obtained the worst performance with a ROC AUC of 0.6179 and a PR AUC of 0.0773. This can be explained by the average distance of each community centroid to other community centroids on the PCA plot (see the third column of Table 7): $Community\ 2$ was the farthest apart from the rest, with an average distance of 2.562.

|  | ROC AUC | PR AUC | average distance to other communities |
|---|---|---|---|
| FL | 0.6895 | 0.1107 | - |
| CBFL Community 1 | 0.7561 | 0.1291 | 1.253 |
| CBFL Community 2 | 0.6179 | 0.0773 | 2.562 |
| CBFL Community 3 | 0.7168 | 0.1321 | 1.420 |
| CBFL Community 4 | 0.7454 | 0.2155 | 2.101 |
| CBFL Community 5 | 0.6967 | 0.1548 | 1.397 |

Table 7: CBFL with 5 communities in the *mortality* prediction task; community comparison in terms of ROC AUCs, PR AUCs, and average distance to other communities, with FL's ROC AUC and PR AUC as a reference

## 4. Discussion

Patients admitted to ICUs come from diverse ethnic and age groups, exhibit various levels of vital sign measurements and illness severity, and receive different diagnoses and treatment [34]. Among these dimensions, CBFL focused primarily on admission diagnoses for patients' unit stay and also on geological locations of hospitals. Via clustering patients of common features into the same community and learning separate models for individual communities, the algorithm converged to higher predictive accuracy in less communication rounds than the baseline FL model in both mortality and stay time prediction tasks. Clustering also made prediction results interpretable: analyzing the distances between communities could help explain why prediction on some examples was more reliable than on others (refer to Table 7 as an example). Moreover, unlike other optimization algorithms for federated learning on non-IID data [26-28] that required a fraction of all data to be shared across the clients, CBFL obviated any degrees of patient data transmission, thereby keeping privacy intact. Any data sent to the server for fitting the clustering model $f_{kmeans}$ was firstly encoded by $f_{encoder}$ and, since the decoder was discarded at the end of training $f_{autoencoder}$ on each client, recovering original *drug features* from encoded representation was nearly impossible.

On the other hand, a limitation of CBFL was that, if $K$ community models were trained on each client, then $K-1$ times more model parameters would be transferred between the clients and the server than those of FL. Such additional communication load would escalate with increasing training samples and communities. Fortunately, experimental results show that CBFL performed the best with five or at most ten communities, but nevertheless there is no guarantee that applications of CBFL on other biomedical datasets will also be the most performant with fewer communities. Future research directions may include optimization of communication load by devising more efficient community-based learning schemes, incorporation of more dimensions other than *drug features* to further boost prediction accuracy, and development of better clustering methods that capture comprehensive characteristics of patients, not only diagnoses and locations but also age, weight, height and more.

**5. Conclusions**

This study presents a novel federated machine learning model CBFL that sought to tackle the challenge of non-IID ICU patient data that complicated decentralized learning, cluster patients into clinically meaningful communities, and optimize performance of predicting mortality and ICU stay time. Our model was evaluated against the baseline FL model on three metrics, namely, ROC AUC that quantifies the likelihood of a model ranking a positive example over a negative one, PR AUC that measures prediction success of a model given datasets with imbalanced labels, and communication rounds that indicate a model's learning speed. Experimental results show that CBFL had predictive accuracy close to that of centralized learning, hence alleviating the non-IID problem, and that it outperformed FL in terms of all three metrics and in every prediction task, whether it be mortality or stay time prediction, and with or without same training/test data distributions. Patient communities formed by CBFL contained different overrepresented diagnoses and seemed to accommodate hospitals from diverse geological locations. In addition, performance difference in communities could be attributed to Euclidean distances on the PCA plot. A last point to make is that, while this study concerned with machine learning on ICU EMRs, CBFL could be extended to other biomedical informatics applications, such as medical image recognition or decision-making on medical planning across multiple healthcare silos with large, distributed, and privacy-sensitive data.

**Acknowledgements**